\documentclass{article}
\usepackage{spconf,amsmath,epsfig}
\usepackage{subfloat}
\usepackage{subfig}
\usepackage{algorithm}
\usepackage[noend]{algorithmic}
\usepackage[citecolor=red]{hyperref}

\let\OLDthebibliography\thebibliography
\renewcommand\thebibliography[1]{
  \OLDthebibliography{#1}
  \setlength{\parskip}{0pt}
  \setlength{\itemsep}{0pt plus 0.3ex}
}

\begin{document}\sloppy

\def\x{{\mathbf x}}
\def\L{{\cal L}}

\title{Peer Learning for Unbiased Scene Graph Generation}
%

\name{Liguang Zhou\textsuperscript{1}, Junjie Hu\textsuperscript{1}, Yuhongze Zhou\textsuperscript{2}, Tin Lun Lam\textsuperscript{1}, Yangsheng Xu\textsuperscript{1}\thanks{Corresponding author: Tin Lun Lam.}}
\address{\textsuperscript{1} AIRS, CUHK-Shenzhen, Shenzhen, China, \textsuperscript{2} McGill University, Montreal, Canada}

\maketitle

\begin{abstract}
Unbiased scene graph generation (USGG) is a challenging task that requires predicting diverse and heavily imbalanced predicates between objects in an image. To address this, we propose a novel framework “peer learning” that uses predicate sampling and consensus voting (PSCV) to encourage multiple peers to learn from each other. Predicate sampling divides the predicate classes into sub-distributions based on frequency, and assigns different peers to handle each sub-distribution or combinations of them. Consensus voting ensembles the peers’ complementary predicate knowledge by emphasizing majority opinion and diminishing minority opinion. Experiments on Visual Genome show that PSCV outperforms previous methods and achieves a new state-of-the-art on SGCls task with \textbf{31.6\%} mean on mR@50/100 and R@50/100.
\end{abstract}

\section{Introduction}
\label{sec:intro}

Scene graph generation has been a topic of increasing interest in the research community, with a growing number of methods proposed to address this problem. The rapid progress in this area has highlighted the potential of scene graph generation for a range of applications, including visual question answering and 3D scene graph generation for robotics. However, most existing approaches to scene graph generation suffer from bias, in which the most commonly predicted predicates are "on" and "has". This bias leads to less informative and less descriptive scene graph predictions, which still have major room for improvement.

In recent years, there has been a shift towards generating scene graphs without bias towards certain predicates in order to improve the quality and informativeness of scene graph predictions. These approaches, known as unbiased scene graph generation (USGG), have the potential to substantially improve the field of scene graph generation. There are few methods to solve this problem using a set of classifiers. For instance, a group of classifiers can be used to incrementally learn USGG from head-to-tail classes, as demonstrated by Dong et al. in their work on stacked USGG \cite{dong2022stacked}. The divide and conquer strategy is proposed with multiple classifiers to encourage learning the different parts of the dataset \cite{han2022divide}. Both of these methods relied on the knowledge distillation method to learn from head to tail. The mixture of experts with a two-level dynamic weighting module is proposed to adjust the weights of each classifier and each predicate prediction for a more diversified expert network for USGG \cite{zhou2022came}. However, all of these methods have not considered combining the opinions of each classifier on predicate prediction in a majority confidence voting manner.

\begin{figure}[t]
	\centering
	\subfloat{
		\begin{minipage}{0.45\textwidth}
			\centering
			\includegraphics[width=8cm]{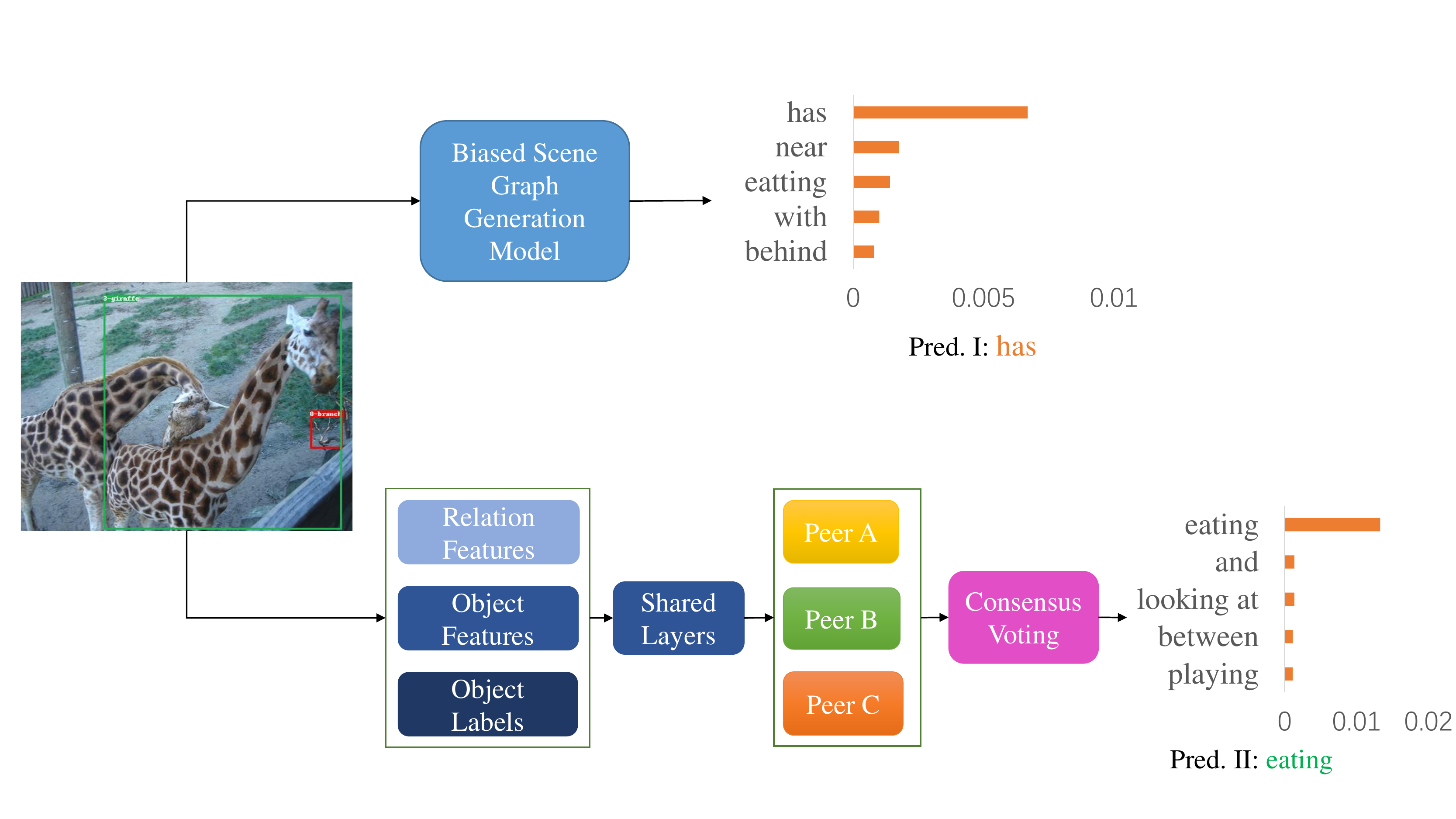}
	\end{minipage}}
	\caption{Differences between our method and conventional USGG methods. a) Conventional methods only train a single model on the Visual Genome dataset, which is inferior in dealing with the intrinsic heavily biased predicate distributions b) Our method divide the long-tailed distribution of predicates into different distributions.  Then, we propose a novel peer learning framework to conquer each sub-distribution individually. Last, the prediction results of each peer are ensembled through a consensus voting strategy.}
    \label{fig:usgg_method}
    \end{figure}

In this paper, we propose a novel peer learning framework that enables multiple peers to learn from each other by exchanging their predictions on predicate labels using majority confidence voting, as illustrated in Fig.~\ref{fig:usgg_method}. Here, the term “peer” refers to a learner that collaborates with other learners to achieve a common goal \cite{mustafa2017learning}. Firstly, we employ predicate sampling to assign different predicate distributions to different peers. This way, each peer can learn different predicate knowledge based on its input predicate distributions. Then, majority confidence voting is proposed to calculate the final prediction by combining the predictions of the peers. This approach allows for leveraged complementary knowledge among peers, thereby improving the accuracy and diversity of their predictions in the long-tailed class imbalance problem.

Our primary contributions are outlined below. 1) We present a predicate sampling strategy for each peer to learn complementary predicate knowledge for USGG. First, the dataset is divided into different sub-distributions initially based on the frequency of predicates (i.e., head, body, and tail classes). These sub-distributions are then grouped together and distributed to different peers to provide them with complimentary knowledge regarding full predicate distributions. 2) We propose a consensus voting strategy that simulates the majority confidence voting process in our society, where the majority opinion with maximum confidence is emphasized, and the minority opinion diminishes to some degree. 3) We have reached \textbf{31.6} mean on the SGCls task using the VCTree baseline, indicating the establishment of a new SOTA.

\section{Related Work}

\subsection{Unbiased Scene Graph Generation}
The topic of USGG has garnered significant attention in recent years, due to its focus on addressing the inherent heavily long-tail distributions of predicates. Various approaches have been proposed to tackle this problem. Energy based loss function  is proposed to efficiently incorporate the structure of scene graphs to alleviate biased scene graph generation \cite{suhail2021energy}. The pixel-level segmentation grounded scene graph generation method is used to generate balanced scene graphs. These method does not fundamentally address the USGG problem \cite{khandelwal2021segmentation}. Additional semantic information or scene graph structures are limited to be associated with USGG. Besides, Chen et al. have proposed using logits adjustment to alleviate biased scene graph generation \cite{chen2022resistance}. Moreover, the Bipartite Graph Network is proposed by Li et al. in \cite{li2021bipartite}. This method uses a relationship confidence estimation module and confidence-aware message propagation module to improve the quality and informativeness of scene graph predictions iteratively. The semantic similarities are considered by exploiting the predicate probability distribution \cite{li2022ppdl}. A noisy label correction strategy is proposed for USGG. This paper raises the concern that previous SGG either favors the head or the tail, which is not comprehensively evaluated. Therefore, the mean is proposed to better evaluate the USGG performance \cite{li2022devil}. However, most of these methods does not address the long-tail distribution essentially by decomposing the datasets according to predicate distributions. In contrast, we propose a novel peer learning framework, with the purpose to interact and collaborate between different peers when acquiring predicate knowledge. 

	\begin{figure*}[t]
	\centering
	\subfloat{
		\begin{minipage}{1\textwidth}
			\centering
			\includegraphics[width=14cm]{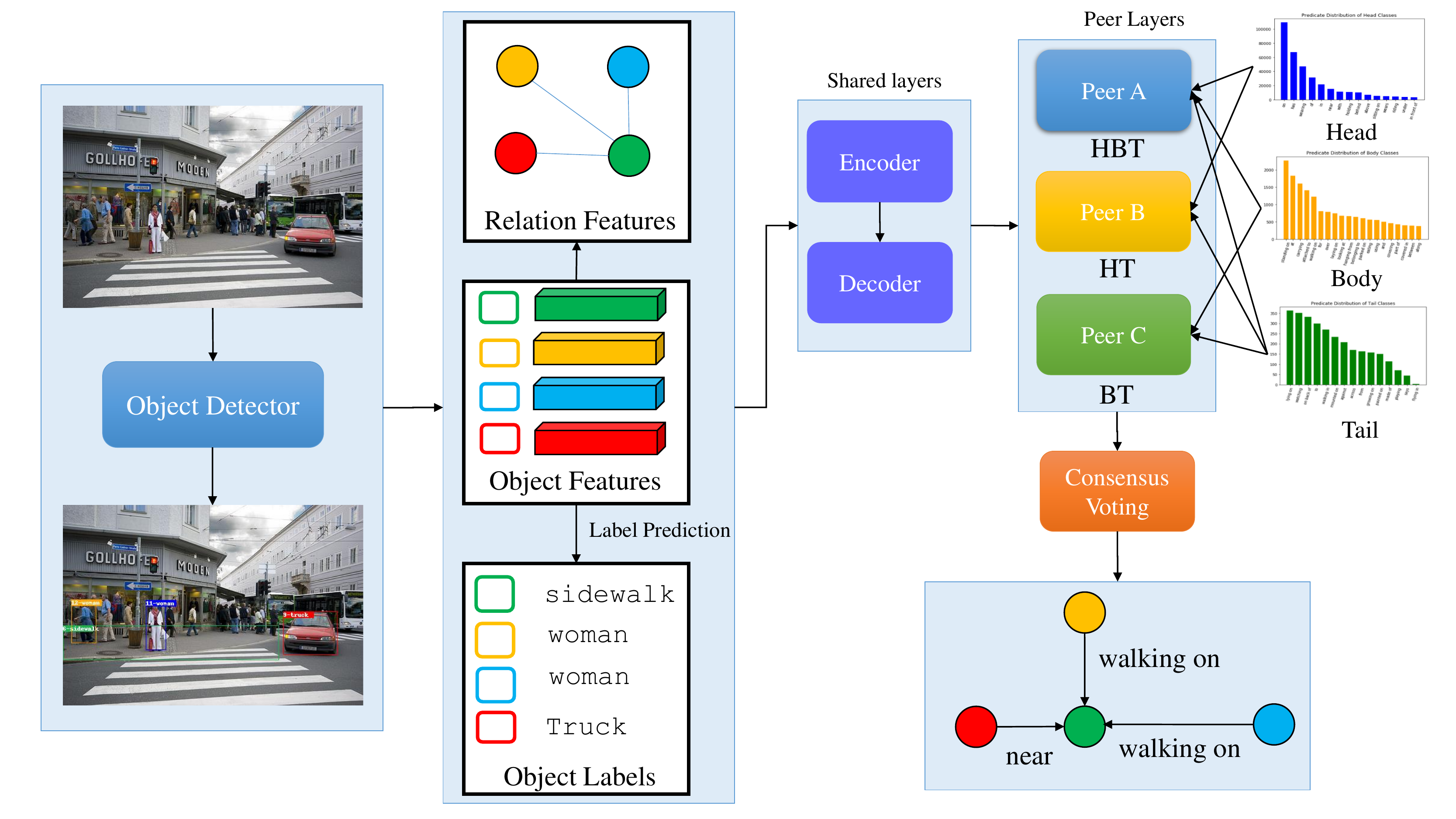}
	\end{minipage}}
	\caption{The proposed peer learning framework for USGG is depicted. We adopt the two-stage scene graph generation method. In the first stage, the object proposals, object labels, and relation features are estimated using the build-in SGG network $\operatorname{P}(B \mid I)$ and  $\operatorname{P}(O \mid B, I)$. In the second stage,  we utilize the built-in encoder/decoder network for relationship information processing. To construct the peer layers to expand the learning capabilities of original build-in scene graph generation methods, we construct multiple peers with various learning materials to build up various peers, denoted as $(P(R_i | B, O, I))$. To obtain USGG results, the opinion of various peers is ensembled through a consensus voting strategy.}
	\label{fig:method}
\end{figure*}

\section{Approach}
Typical scene graph generation methods have one predictor, $\operatorname{P}(G \mid I)$, with one dominant peer to decide the predicate between the subject to object, which could be biased toward some head classes while neglecting the body and tail classes.

\begin{equation}
	\operatorname{P}(G \mid I)=\operatorname{P}(B \mid I) \operatorname{P}(O \mid B, I) \operatorname{P}(R \mid B, O, I),
\end{equation}

One peer is quite limited, which eventually leads to biased scene graph generation due to the limited knowledge learned and retained. To alleviate this situation, we present a framework, doubted as peer learning, with predicate sampling and consensus voting (PSCV) strategy, to address the long-tailed distribution problem in a divide-and-vote manner. To achieve this, the first step is to divide the heavily long-tailed problem into more balanced sub-distributions within a smaller group of predicates. Thus, we utilize predicate sampling to sample the subset of the Visual Genome dataset into head, body, and tail classes. Then, we pack the different subsets as learning material for one single peer to learn. To ensure diversified peers, the learning material acquired by each peer is varied and diversified. Specifically, we promise a major peer to learn the whole perspective of datasets, while the rest peers to learn the complementary knowledge as a supplement to fill the gap between the body and tail classes to head classes.

\subsection{Predicate Sampling}
In this section, we introduce peer learning for USGG. To address this problem, we present a peer learning framework with various peers, with different learning materials for each peer, as depicted in Fig.~\ref{fig:method}. In this case, we obtain $\operatorname{P}(G_i \mid I)$,

\begin{equation}
	\operatorname{P}(G_i \mid I)=\operatorname{P}(B \mid I) \operatorname{P}(O \mid B, I) \operatorname{P}(R_i \mid B, O, I), i \in [1,n],
\end{equation}

where $i$ stand for the $i-{th}$ peer of the whole ensemble network, and n stands for the number of peers in the ensemble network. Based on the assumption that, if we feed each peer with different and complementary learning material, they can learn different and complementary information. To achieve this, we first divide the heavily long-tailed distributions into more balanced sub-distributions for each peer to learn, then each peer can have a better understanding of the sub-distributions, and their ensembles can have a better understanding of the entire heavily long-tailed distributions. This approach can help to reduce the effects of imbalanced data and enable the algorithm to learn more effectively from the sub-datasets. However, it is important to carefully design the sub-datasets in a way that ensures they still contain a representative sample of the overall data distribution. Otherwise, the resulting ensembles may not accurately reflect the entire dataset. Additionally, it is important to consider the trade-offs between the improved performance on the sub-datasets and the loss of information that comes with dividing the data in this way. We put this assumption into practice. First, we divide the network with various peers and we have $\operatorname{P}(G_i \mid I)$. The next stage is to feed each peer with different complementary targets for USGG. 

\subsection{Consensus Voting}

The opinions of each peer towards predicate prediction are different based on their different knowledge sources and combinations of various predicate distributions. To achieve a consensus among peers with such situation, we propose a novel consensus voting strategy that based on the voting predicate and corresponding confidence scores of each peer and performs a majority confidence voting process. Algorithm~\ref{alg:cv} shows the details of this strategy.

\section{Experimental Results}
\begin{table*}[t]
	\small
	\renewcommand{\arraystretch}{1}
	\caption{SGG performs in percentage (\%) of various methods in the Visual Genome dataset. Mean Recall (mR) is reported over three tasks.}
	\label{tab:comparsion}
	\begin{center}
		\begin{tabular}{c|ccc|ccc|ccc} 
			\hline
			& \multicolumn{3}{c}{PredCls} & \multicolumn{3}{c}{SGCls} & \multicolumn{3}{c}{SGGen}   \\ \hline
			Model		  & mR@20 & mR@50 & mR@100 & mR@20 & mR@50 & mR@100 & mR@20 & mR@50  & mR@100	\\ \hline
			IMP+ \cite{xu2017scene,chen2019knowledge}            & - & 9.8 & 10.5 & - & 5.8 & 6.0 & - & 3.8 & 4.8  					\\
			FREQ \cite{zellers2018neural,tang2019learning}	   & 8.3 & 13.0 & 16.0 & 5.1 & 7.2 & 8.5 & 4.5 & 6.1 & 7.1  			\\ 
			KERN \cite{chen2019knowledge}
			& - & 17.7 & 19.2 & - & 9.4 &  10.0 & - & 6.4 & 7.3 \\
			GPS-Net \cite{lin2020gps}
			& 17.4 & 21.3 & 22.8 & 10.0 & 11.8 & 12.6 & 6.9 & 8.7 & 9.8 \\
			GB-Net \cite{zareian2020bridging}
			& - & 22.1 & 24.0 & - & 12.7 & 13.4 & - & 7.1 & 8.5  \\ 	
			Seq2Seq - RL \cite{lu2021context}
			& 21.3 & 26.1 & 30.5 & 11.9 & 14.7 & 16.2 & 7.5 & 9.6 & 12.1	\\ \hline
			MOTIFS \cite{zellers2018neural}	   & 12.6 & 16.1 & 17.4 & 6.5 & 8.0 & 8.5 & 5.3 & 7.3 & 8.6  \\
			MOTIFS-Focal$\star$				   & 12.2 & 16.2 & 18.2 & 7.0 & 8.9 & 9.7 & 4.5 & 6.4 & 7.7  \\ 
			MOTIFS-LDAM$\star$				   & 12.6 & 15.9 & 17.2 & 7.2 & 8.9 & 9.5 & 5.2 & 7.0 & 8.2 	\\ 
			MOTIFS-TDE (SUM)$\star$ \cite{tang2020unbiased}           & 16.3 & 22.9 & 26.9 & 10.2 & 13.7 & 15.6 & 6.7 & 9.1 & 10.9 		\\
			MOTIFS-EBM \cite{suhail2021energy}     			   & 14.2 & 18.0 & 19.5 & 8.2 & 10.2 & 11.0 & 5.7 & 7.7 & 9.3 			\\ 
			MOTIFS-Seg \cite{khandelwal2021segmentation}   	& 14.5 & 18.5 & 20.2 & 8.9 & 11.2 & 12.1 & 6.4 & 8.3 & 9.2  		\\ 
            MOTIFS-TDE-IL \cite{goel2022not}
            & 21.3 & 27.1 & 29.7 & 11.3 & 14.3 & 15.7 & 8.4 & 10.4 & 11.8 \\
   \hline
			MOTIFS-PSCV              & \textbf{28.1} & \textbf{33.2} & \textbf{35.2} & 
			\textbf{17.1} & \textbf{20.8} & \textbf{21.9} & 
			{6.4}  & 
			{10.1}  & 
			\textbf{13.9}  \\ \hline
			VCTree \cite{tang2019learning}	   & 13.7 & 17.4 &	19.0 & 8.1 & 9.9 & 10.6 & 5.3 & 6.9 & 7.9 			\\ 
			VCTree-Focal$\star$				   & 7.3  & 10.9 & 13.3 & 8.7 & 11.0& 11.9 & 4.3 & 6.3 & 7.6 		\\ 
			VCTree-LDAM	$\star$				   & 7.5  & 10.7 & 12.8 & 7.9 & 9.5 & 10.1 & 4.0 & 5.3 & 6.1  			\\ 
			VCTree-TDE(SUM)* \cite{tang2020unbiased}			& 19.5 & 26.2 & 29.8 & 10.5 & 15.0 & 17.4 & 6.9 & 9.6 & 11.5 	\\ 
			VCTree-EBM \cite{suhail2021energy}				 & 14.2 & 18.2 & 19.7 & 10.4 & 12.5 & 13.5 & 5.7 & 7.7 & 9.1   		\\
   VCTree-Seg\cite{khandelwal2021segmentation}               &  15.0 & 19.2 & 21.1 &  9.3 & 11.6 & 12.3 & 6.3 & 8.1 & 9.0  	\\ 
   VCTree-IL \cite{goel2022not} 
            & 18.0 & 21.7 & 23.1 & 11.9 & 14.1 & 15.2 & 7.1 & 8.2 & 8.7 \\ \hline
			VCTree-PSCV
			& \textbf{30.2} & \textbf{34.8} & \textbf{37.2} & \textbf{22.1} & \textbf{25.8} &  \textbf{27.1} & 6.5 & \textbf{9.9} & \textbf{13.3} \\ \hline 
			BGNN \cite{li2021bipartite}
			& - & 30.4 & 32.9 & - & 14.3 & 16.5 & - & 10.7 & 12.6 \\ 
			BGNN-PSCV &  \textbf{26.1} & \textbf{33.2} & \textbf{36.1} & \textbf{14.2} & \textbf{17.8} & \textbf{19.5} & \textbf{8.6} & \textbf{12.3} & \textbf{15.2} \\  \hline
			Transformer & 13.9 &  17.7 & 19.1 & 8.4 & 10.4 & 11.0 & 6.0 & 8.2 & 9.6  \\ 
			Transformer-PSCV & \textbf{30.2} & \textbf{34.8} & \textbf{36.4} & \textbf{18.6} & \textbf{21.3} & \textbf{22.2} & \textbf{7.3} & \textbf{11.1} & \textbf{14.5} 
			\\ \hline				
		\end{tabular}
	\end{center}
\end{table*}

\begin{algorithm}[!h]
	\caption{\label{alg:cv} Consensus voting strategy for all peers.} 
	\begin{algorithmic}[1]
		\REQUIRE The predicate labels are defined as $P_l\in R^{n \times 1}$ and their confidence scores are defined as  $P_s \in R^{n \times 1}$ ;
		\STATE $v_n$ is the voting number, max\_score = 0.0, max\_label = 0
		\STATE Construct the voting number dictionary $D_{vn}$ of all predicate labels, e.g, if 2 peers vote for predicate label 2  and 1 peer votes for predicate label 5, then the $D_{vn}$=\{2:2, 5:1\}
		\FOR{$p_l$, $v_n$ in $D_{vn}.items()$}
		\IF{$v_n$==len($P_l$)}
		\STATE return max($P_s$),$p_l$
		\ELSIF{$v_n$ $>$ 1}
		\FOR{i in range(len($P_l$)) }
		\IF{ $p_l$==$P_l[i]$ }
		\STATE $v_s == P_s[i]$
		\IF{$v_s$ $>$ max\_score}
		\STATE max\_score = $v_s$
		\STATE max\_label = $P_l[i]$
		\ENDIF
		\ENDIF
		\ENDFOR
		\ELSIF{$v_n$==1}
		\STATE $v_s$ = $P_s[P_l.index(p_l)]$
		\IF{ $v_s$ $>$ max\_score}
		\STATE max\_score = $v_s$, max\_label = $p_l$ 
		\ENDIF
		\ENDIF
		\ENDFOR
		\STATE return max\_score, max\_label
	\end{algorithmic}
\end{algorithm}

\subsection{Loss function}
The loss function of peer learning framework is composed of three peers, which are summed together as a total loss $L_{pl}$ for USGG. In the experiment, $\alpha$ is configured as 1 by default. The $L_{Cls_i}$ is configured as LDAM, class balanced loss, cross entropy loss, focal loss \cite{zhou2022came}. 

\begin{equation}
\mathcal{L}_{\mathrm{pl}}(x, y)=\sum_{i=1}^n\left(\alpha_i \mathcal{L}_{\mathrm{Cls_i}}\left(x, y_i ; \theta_i\right)\right).
\end{equation}

The $y_i$ is configured with three different distributions as learning targets. To achieve this, we split the dataset into different sub-distributions, i.e., the head, body, and tail distributions independently. Then, we consider adding different distributions to different peers as their  targets for the learning of USGG. For instance, in the baseline settings, the head, body, and tail classes are assigned to three peers independently ($H\_B\_T$). In addition to the baseline settings, we intend to make first peer with an entire overview of the dataset, therefore, the target of the first peer is with head, body, and tail classes, while the 2nd and 3rd peers can be fed with combinations of these sub-distributions, such as body for 2nd peer and tail 3rd peer ($HBT\_B\_T$), or head, body for 2nd, body and tail for 3rd peer ($HBT\_HB\_BT$).

\begin{table}[ht]
	\scriptsize
	\centering
	\caption{Ablation, mR@20/50/100 and R@20/50/100 for SGCls is reported. MOTIFS \cite{zellers2018neural} is the base architecture for all methods.}
	\label{tab:ablation_mot}
	\begin{tabular}{ll|ccc}
		\hline
		& & \multicolumn{2}{c}{SGCls}      \\ \hline
		Model      & Config    & mR@50/100 & R@50/100  & mean \\ \hline
		MOTIFS-PSCV & HBT\_T &  19.7 / 20.6 &  25.0 / 25.3  & 22.65 \\  
		MOTIFS-PSCV & HBT\_B &  20.2 / 20.9 &  25.0 / 25.3  & 22.85 \\ 
		MOTIFS-PSCV & HBT\_H\_B\_T &  20.0 / 20.6 & 25.9 / 26.2 & 23.20 \\  
		MOTIFS-PSCV & HBT\_HB\_BT\_HT &  20.0 / 20.7 &  25.9 / 26.2 & 23.20 \\  
		MOTIFS-PSCV & H\_B\_T &  15.2 / 16.0 &  35.0 / 35.8  & 25.50  \\
		MOTIFS-PSCV & HBT\_B\_T  & 21.0 / 22.1 & 31.5 / 32.4 & 26.75  \\
		MOTIFS-PSCV & HBT\_BT\_T   &  20.9 / 22.1 &  31.0 / 32.0 & 26.50  \\ 
		MOTIFS-PSCV & HB\_HT\_BT  & 16.7 / 17.7 &  33.0 / 34.1  & 25.38  \\
		MOTIFS-PSCV & HBT\_HT\_BT &  20.8 / 21.9 &  32.5 / 33.6  & 27.20  \\ 
		\hline \hline
	\end{tabular}
        \end{table}
        
\subsection{Visual Genome}
Visual Genome is a large-scale dataset for scene graph generation and other visual reasoning tasks. It consists of images and associated annotations, including bounding boxes and labels for objects, relationships between objects, and attributes of objects and relationships. Visual Genome dataset is widely used in research on scene graph generation. Following the settings \cite{zhou2022came}, dataset is split into training set with 62,723 images, validation set with 5,000 images, and test set with 26,646 images. 

\subsection{Evaluation Metrics}
The evaluation of USGG mainly contains three tasks, including 1) predicate classification (PredCls), which involves identifying the type of predicate between two entities  $\ <subject, predicate, object\ >$. 2) scene graph classification (SGCls), which involves assigning a class label to each object and the relationship between each object pair. 3) scene graph generation (SGGen), identifying the objects and their relationships in the image and constructing a scene graph that accurately represents the content of the image. 
To measure the performance of USGG, three metrics are used, namely R@K, mR@K, and Mean. The Top-k recall (R@K) is used to calculate the proportion of relevant predicates in the top K positions of the predicate predictions, which is usually set as 20, 50, and 100 \cite{zellers2018neural, tang2019learning}. The mean recall (mR@K) evaluates the recall of each predicate category, and then averages the recalls over all predicates. Mean is a balanced metric that takes into account both R@K and mR@K, thereby providing a more comprehensive assessment of the model performance across all predicates \cite{li2022devil}.

\subsection{Implementation Details}

\subsubsection{\textbf{Object Detection}}
The framework of object detection is based on a two-stage object detector, Faster RCNN \cite{ren2015faster}, which is pretrained before the training of scene graph generation.

\subsubsection{\textbf{PSCV}}
The proposed method is evaluated in two most popular baselines including Motifs, and VCTree. Besides, we have added the transformer and BGNN baseline for further comparison. The experiments have been conducted on number of peers ranging from 2 to 4.

\begin{table}[ht]
	\scriptsize
	\centering
	\caption{Ablation, mR@20/50/100 and R@20/50/100 for SGCls is reported. VCTree \cite{tang2019learning} is the base architecture for all methods.}
	\label{tab:ablation_vc}
	\begin{tabular}{ll|ccc}
		\hline
		&  & \multicolumn{2}{c}{SGCls}    \\ \hline
		Model      & Config            & mR@50/100 & R@50/100 & mean  \\ \hline
		VCTree-PSCV & H\_B\_T &   19.3 / 20.9 &  39.6 / 40.5     & 30.08 \\
		VCTree-PSCV & HBT\_B\_T & 25.8 / 27.2 & 36.1 / 37.3     & 31.60 \\
		VCTree-PSCV & HBT\_BT\_T  & 25.8 / 27.1 &  35.8 / 37.1   & 31.45 \\ 
		VCTree-PSCV & HB\_HT\_BT &  22.9 / 24.3  & 38.5 / 39.7   & 31.35 \\ 
		VCTree-PSCV & HBT\_HT\_BT  &  25.4 / 27.1 &  34.2 / 35.6 & 30.58 \\
		\hline \hline 
	\end{tabular}
\end{table}

\subsection{Comparison with other state-of-the-art methods}
\begin{figure}[t]
	\centering
	\subfloat{
		\begin{minipage}{0.45\textwidth}
			\centering
			\includegraphics[width=6cm]{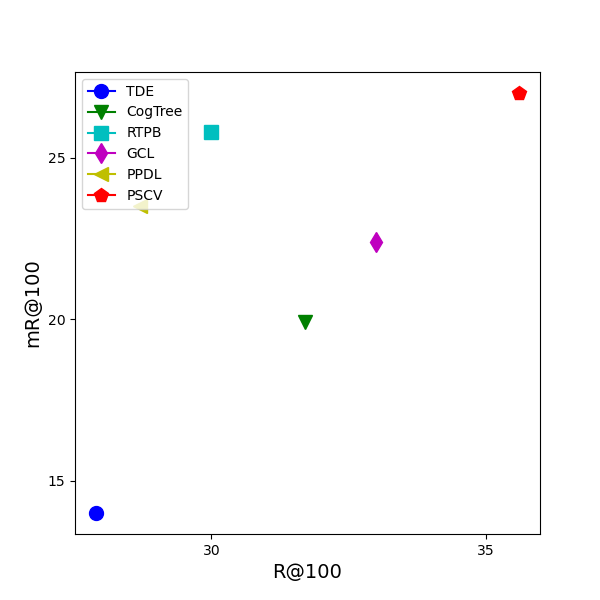}
	\end{minipage}}
	\caption{The comparison between PSCV and current state-of-the-art USGG methods based on VCTree architecture is depicted in the figure on the R@100 and mR@100 of the SGCls task. The result show PSCV has significantly  improved on both mR@100 and R@100, reaching a mean of \textbf{32.3} on USGG performance, outperforming GCL by about 4.6.}
	\label{fig:usgg_compare}
	\end{figure}

To demonstrate the effectiveness of the proposed PSCV method, we compare it against state-of-the-art methods, including energy-based models (EBMs), segmentation-grounded models (Segs), informative labels \cite{goel2022not}, and Bipartite Graph Neural Networks (BGNNs) \cite{li2021bipartite}. Results of the comparison are shown in Tab.~\ref{tab:comparsion}, which shows that our method delivers the best performance compared with the other methods. MOTIFS-PSCV, VCTree-PSCV, and Transformer-PSCV methods have achieved twice as greater mean recall compared with MOTIFS, VCTree, and Transformer, demonstrating that our proposed model is model-agnostic and could be applied to different models. Additionally, our method has achieved \textbf{27.1} mR@100 with the VCTree framework on the SGCls task, which is nearly double the performance compared with VCTree-Seg and VCTree-EBM. This indicates that the proposed PSCV algorithm effectively facilitates multiple peers to understand the different predicate distributions, leading to more balanced predicate predictions. Fig.~\ref{fig:usgg_compare} shows that our proposed method has achieved the best performance trade-off among the current state-of-the-art approaches in the R@100 and mR@100 metrics. Unlike TDE \cite{tang2020unbiased}, CogTree \cite{yu2020cogtree}, RTPB \cite{chen2022resistance}, Group Collaborative Learning (GCL) \cite{dong2022stacked}, and PPDL \cite{li2022ppdl}, which strive to improve performance in terms of mean Recall (which favors head classes) but overlook performance in terms of Recall (which favors tail classes), our method has achieved better performance in both head and tail classes, thus achieving a new state-of-the-art in the SGCls task, with \textbf{35.6} R@100 and \textbf{27.0} mR@100.

\subsection{Ablation Study}

\begin{table}[t]
	\scriptsize
	\centering
	\caption{The SGG performance in SGCls settings. We show the mR@100, and performance on the head, body, and tail classes and its variance based on the VCTree framework.}
	\label{tab:sggcls}
	\begin{tabular}{l|l|llllll}
		\hline
		& mR@100   & body & var  & tail &  var  \\ \hline
		LDAM  & 11.2          & 3.79 & 8.68 & 0.98 & 13.7 \\
		FOCAL & 11.8          & 5.54 & 10.4 & 0.76 & 10.7 \\
		TDE   & 17.4         & 19.8 & 19.2 & 1.40 & 10.4 \\ \hline
		PSCV  & \textbf{27.1}  & \textbf{27.7} & \textbf{8.92} & \textbf{27.4} & \textbf{2.69} \\ \hline \hline
	\end{tabular}
\end{table}

In the ablation study, the performance of USGG was evaluated by feeding different combinations of distributions to various peers. As Tab.~\ref{tab:ablation_mot} showed, the best performance was achieved when the number of peers was set to three. Thus, the typical setting for the number of peers was three. Moreover, it was observed that the scene graph generation network with independent head, body, and tail distributions established a baseline performance, which leaves a margin for improvement.
As the entire view of data distributions have been added to the first peer ($ HBT\_B\_T$), the performance has gained greatly. The different combinations of distributions are used during training, such as $ H\_B\_T$ and $ HBT\_HT\_BT$. The best performance is 27.2 mean with $HBT \_HT\_BT$ settings. As shown in Tab.~\ref{tab:ablation_vc} with VCTree backbone, the phenomena are similar to MOTIFS backbone. We have showed a mean of \textbf{31.60} on the VCTree baseline with $ HBT\_B\_T$ settings. Furthermore, we have compared our proposed method with different debiasing  strategies as shown in Tab.~\ref{tab:sggcls}. It shows the performance of different debiasing strategies on the mean recalls of entire dataset and on the body and tail classes, respectively. In comparison with others, we have achieved \textbf{27.1} mR@100, where the body and tail classes are significantly enhanced through the proposed PSCV algorithm. For instance, we have achieved \textbf{27.7} and \textbf{27.4} mean recalls on the body, and tail classes. Specifically, in the tail classes, we are \textbf{18.57} times more than the TDE \cite{tang2020unbiased} method.

\section{Conclusion}
The biased scene graph generation problem is based on the intrinsic long-tailed predicate distributions, which are heavy and hard to handle with a single network. In this paper, we present a new framework, dubbed peer learning, to deal with the USGG problem in a divide-and-vote manner. The long-tailed problem is solved with three steps. First, we divide the heavy distribution into a subset of more balanced distribution groups, including head, body, and tail classes. Then, these distributions are combined and fed to different peers to increase the diversity  and discriminability of each peer. The results of these peers are ensembled through a consensus voting strategy to merge the opinions by enhancing the majority's opinion and depressing the minority's opinion. The experiment on the Visual Genome dataset shows the efficacy of the proposed PSCV model, which is well-suited the USGG problem.



\small
\bibliographystyle{IEEEbib}
\bibliography{Ref}

\end{document}